\documentclass[letterpaper, 10 pt, conference]{ieeeconf}  

\IEEEoverridecommandlockouts                              

\usepackage{cite}
\usepackage{svg}
\usepackage{amsmath,amssymb,amsfonts}
\usepackage{algorithm}
\usepackage{algpseudocode}
\usepackage{graphicx}
\usepackage{textcomp}
\usepackage{xcolor}
\usepackage{dsfont}
\usepackage{subfigure}
\usepackage{booktabs}
\usepackage{bbm}
\usepackage{url}
\usepackage{svg}
\usepackage{multirow}
\usepackage{multicol}
\usepackage{rotating}
\usepackage{color}
\usepackage{tikz}
\usepackage{bigstrut}
\usepackage{array}
\usepackage{tabularx}
\usepackage{siunitx}
\usepackage{amssymb}
\usepackage{pifont}

\usepackage[normalem]{ulem}
\useunder{\uline}{\ul}{}
\usepackage{marvosym}

\makeatletter
\let\NAT@parse\undefined
\makeatother
\usepackage{hyperref}
\usepackage[nameinlink,noabbrev]{cleveref}

\definecolor{opengt}{rgb}{0.863, 0.929, 0.969}
\definecolor{openpred}{rgb}{0.836, 0.836, 0.836}
\definecolor{quick-module}{rgb}{0.953, 0.718, 0.596}
\definecolor{slow-module}{rgb}{0.820, 0.820, 0.820}

\definecolor{scene}{rgb}{0.984, 0.890, 0.839}
\definecolor{object}{rgb}{0.792, 0.933, 0.984}
\definecolor{action}{rgb}{0.850, 0.949, 0.816}

\title{\LARGE \bf
Dual-AEB: Synergizing Rule-Based and Multimodal Large Language Models for Effective Emergency Braking
}

\author{Wei Zhang$^{1,3,*}$, Pengfei Li$^{1,*}$, Junli Wang$^{1,4}$, Bingchuan Sun$^{2}$, Qihao Jin$^{5}$, \\ 
Guangjun Bao$^{2}$, Shibo Rui$^{2}$, Yang Yu$^{2}$, Wenchao Ding$^{5}$, Peng Li$^{1}$\textsuperscript{\Letter} and Yilun Chen$^{1}$\textsuperscript{\Letter}
\thanks{$^{1}$ Institute for AI Industry Research (AIR), Tsinghua University, China,
        \{lipeng, chenyilun\}@air.tsinghua.edu.cn, li-pf22@mails.tsinghua.edu.cn.}%
\thanks{$^{2}$ Lenovo Research.}%
\thanks{$^{3}$ Harbin Institute of Technology, China, hitwizard@outlook.com.}%
\thanks{$^{4}$ University of Chinese Academy of Sciences, China,
        wangjunli2022@ia.ac.cn.}%
\thanks{$^{5}$ Academy for Engineering and Technology, Fudan University, China,
        qhjin24@m.fudan.edu.cn, dingwenchao@fudan.edu.cn.}%
\thanks{$^{*}$ These authors have contributed equally to this work.}%
}
\begin{document}

\maketitle
\thispagestyle{empty}
\pagestyle{empty}

\begin{abstract}

Automatic Emergency Braking (AEB) systems are a crucial component in ensuring the safety of passengers in autonomous vehicles. Conventional AEB systems primarily rely on closed-set perception modules to recognize traffic conditions and assess collision risks. To enhance the adaptability of AEB systems in open scenarios, we propose Dual-AEB, a system combines an advanced multimodal large language model (MLLM) for comprehensive scene understanding and a conventional rule-based rapid AEB to ensure quick response times. To the best of our knowledge, Dual-AEB is the first method to incorporate MLLMs within AEB systems. Through extensive experimentation, we have validated the effectiveness of our method. Codes will be publicly available at \url{https://github.com/ChipsICU/Dual-AEB}.

\end{abstract}


\section{Introduction}
\label{intro}

The Autonomous Emergency Braking (AEB) system is a critical safety feature in autonomous vehicles, designed to mitigate or prevent collisions by automatically activating the brakes when a potential collision is detected\cite{accidents_4}. Numerous studies\cite{accidents_4, cicchino2017effectiveness, jeong2013methodology, fildes2015effectiveness, isaksson2016evaluation} have demonstrated the effectiveness of AEB systems, with reductions in rear-end collisions ranging from $\textbf{25\%}$ to $\textbf{50\%}$.

Conventionally, AEB systems can be roughly categorized into two types: decision-making-only methods\cite{traffic_joiner_3,traffic_joiner_4,traffic_joiner_5,traffic_joiner_6, modified_ttc_1, modified_ttc_2, modified_ttc_3, learning_method_1,learning_method_3, DTTCNet} and end-to-end methods\cite{learning_method_5, learning_method_6}. 
Decision-making-only methods use  perception results of predefined perception categories (e.g., people, cars, bicycles) and apply rule-based techniques\cite{modified_ttc_1, modified_ttc_2, modified_ttc_3, sharan2023llmassist} or deep reinforcement learning\cite{learning_method_1,learning_method_3} for braking decisions. End-to-end methods\cite{learning_method_5, learning_method_6}, meanwhile, process raw sensory data directly to inform AEB decisions, allowing the system to benefit from comprehensive sensory inputs. These methods generally ensure safety in most driving scenarios.

\begin{figure}[tpb]
\centerline{\includegraphics[width=0.5\textwidth]{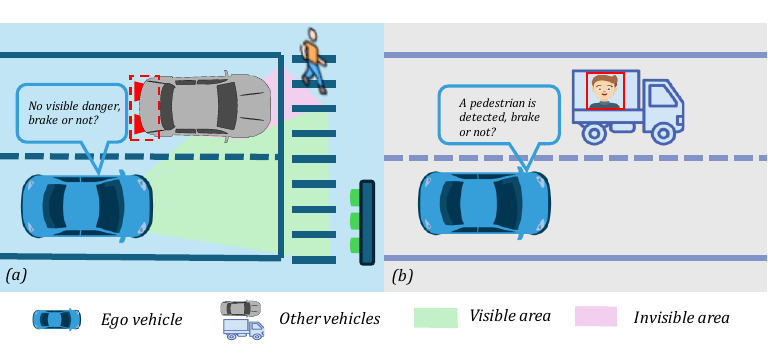}}
\caption{Conventional AEB systems tend to fail in these situations: \textbf{(a)} when early detection of a pedestrian is required to brake in advance and avoid danger, and \textbf{(b)} when incorrect perception triggers AEB unnecessarily. Such scenarios are challenging for conventional AEB methods.}
\label{fig:teaser}
\end{figure}

However, their ability to handle complex driving situations is limited due to a lack of comprehensive scene understanding. For example, in Fig.~\ref{fig:teaser}~(a), the scene describes a pedestrian positioned in the ego vehicle's blind spot, intending to cross at a green-light intersection. Typically, decision-making-only methods would not activate braking in this scenario due to the absence of pedestrian perception information, making it impossible to predict an impending collision. Similarly, while end-to-end methods process raw sensory data, they often lack the reasoning capacity to interpret indirect cues—such as the illuminated brake lights on the vehicle to the left of the ego vehicle—that may indicate a potential hazard ahead. In Fig.~\ref{fig:teaser}~(b), a truck with a facial advertisement is driving on the ego vehicle's left. Both decision-making-only and end-to-end methods may misinterpret the advertisement as a pedestrian, potentially triggering the AEB system and causing unnecessary braking. A truly effective AEB system should incorporate comprehensive scene understanding, enabling it to differentiate between real hazards and non-threatening elements, thereby ensuring appropriate braking responses.

To address these challenges, we propose the \textbf{Dual-AEB} system, which offers the following main advantages: (1) \textbf{Comprehensive Scene Understanding}: The Dual-AEB system integrates advanced Multimodal Large Language Models (MLLMs) to achieve a deep understanding of the driving environment. By processing comprehensive data—including environmental conditions, critical perception information, and ego-vehicle states—MLLMs enhance overall situational awareness while reducing the risk of false positives and missed detections. (2) \textbf{Optimized Response Time}: The Dual-AEB system leverages the strengths of both the conventional AEB module and MLLM components. The conventional AEB ensures a quick initial response to imminent threats, while the MLLM component provides detailed analyses in complex scenarios. This synergistic approach minimizes response time and maximizes accuracy during critical moments. (3) \textbf{Flexible Modular Design}: The Dual-AEB system's modular architecture facilitates seamless upgrades and component replacements as technology advances. This feature guarantees long-term efficacy and continuous improvement, preparing the system to meet future challenges.

To summarize, our contributions are as follows:

\begin{itemize}
\item[$\bullet$] We present Dual-AEB, the first work that integrates MLLMs to enhance conventional AEB systems by leveraging their comprehensive scene understanding to improve braking decisions.
\item[$\bullet$] Our method is validated through extensive experiments on both open-loop and closed-loop benchmarks, demonstrating its effectiveness.
\item[$\bullet$] Qualitative analysis on our in-house real-world scenario dataset further confirms the practicality of deploying this system.
\end{itemize}

\section{Related Work}
\subsection{Autonomous Emergency Braking (AEB)}
AEB systems are essential for vehicle safety, as it autonomously detects risks and activates brakes to mitigate or avoid collisions, significantly reducing traffic accident rates\cite{accidents_1,accidents_2,accidents_3,accidents_4, cicchino2017effectiveness, jeong2013methodology, fildes2015effectiveness, isaksson2016evaluation}. Over time, AEB systems have evolved to utilize either decision-making-only or end-to-end methods, ensuring safety in general scenarios. 

\textbf{Decision-Making-Only Methods.} These methods typically rely on a limited set of closed-set perception results, such as detecting pedestrians, vehicles, and bicycles, to determine the necessity of braking actions. These decisions are often based on metrics like  Time To Collision (TTC)\cite{modified_ttc_1, modified_ttc_2, modified_ttc_3, sharan2023llmassist} or are designed using control algorithms\cite{control_algorithm_1,control_algorithm_2,control_algorithm_3,control_algorithm_4}, and sometimes involve learning-based approaches\cite{learning_method_1,learning_method_3, DTTCNet}. While these methods are straightforward and computationally efficient, they suffer from significant limitations in complex, dynamic environments\cite{traffic_joiner_1,traffic_joiner_2,traffic_joiner_3,traffic_joiner_4,traffic_joiner_5,traffic_joiner_6}. Relying on a predefined closed-set of objects can result in the omission of vital environmental information, potentially leading to the failure of the AEB in critical situations.

\textbf{End-to-End Methods.} End-to-end methods bypass traditional decision-making pipelines by directly using raw perception data for AEB decisions\cite{learning_method_2,learning_method_3,learning_method_4,learning_method_5,learning_method_6,Wu2021RealTimeVA}. They offer flexibility and can continuously improve with more data\cite{9675885}, enabling the detection and response to hazards that rule-based systems might miss. However, these approaches face challenges, including the need for large labeled datasets, inconsistent performance on unseen scenarios\cite{10614862,10.1145/3209889.3209897}, susceptibility to overfitting, and opaque decision-making processes\cite{duede2023deep}, which hinder their reliability in critical and complex situations.

Overall, both decision-making-only and end-to-end methods face challenges in handling complex driving scenarios. Decision-making-only methods rely on predefined perception categories, limiting their ability to respond to unexpected elements. End-to-end AEB systems, while processing raw sensory data, often struggle with reasoning through complex relationships in the scene. To provide effective braking decisions in complex driving scenarios, a comprehensive understanding of the scene is required for AEB systems.

\subsection{Multimodal Large Language Models (MLLMs)}
The advent of large models such as ChatGPT\cite{gpt3,Mirsoft} and Gemini has brought us closer to achieving trustworthy autonomous driving\cite{gptdriver,lmdrive,languagempc,drivelikehuman,10611018, tian2024drivevlmconvergenceautonomousdriving} and robotics\cite{robot_1, robot_2, robot_3, robot_4, robot_5, robot_6, robot_7, robot_8}.
Works like GPT-Driver\cite{gptdriver,lmdrive,languagempc,drivelikehuman,10611018,zheng2024planagentmultimodallargelanguage} have demonstrated superior performance over previous methods through prompt-tuning and fine-tuning techniques in autonomous driving benchmarks. DriveVLM\cite{tian2024drivevlmconvergenceautonomousdriving} introduces a dual system design, akin to the human brain’s slow and fast thinking processes, which efficiently adapts to varying complexities in driving scenarios. This innovative approach helps end-to-end autonomous driving models address corner cases and enhances the overall system's performance ceiling, closely aligning with the focus of our work. Inspired by these works, we integrate MLLMs into AEB systems, aiming to enhance their ability for comprehensive scene understanding and, in turn, provide more effective braking decisions.


\section{Dual-AEB}

\begin{figure*}[t]
\centerline{\includegraphics[width=0.98\textwidth]{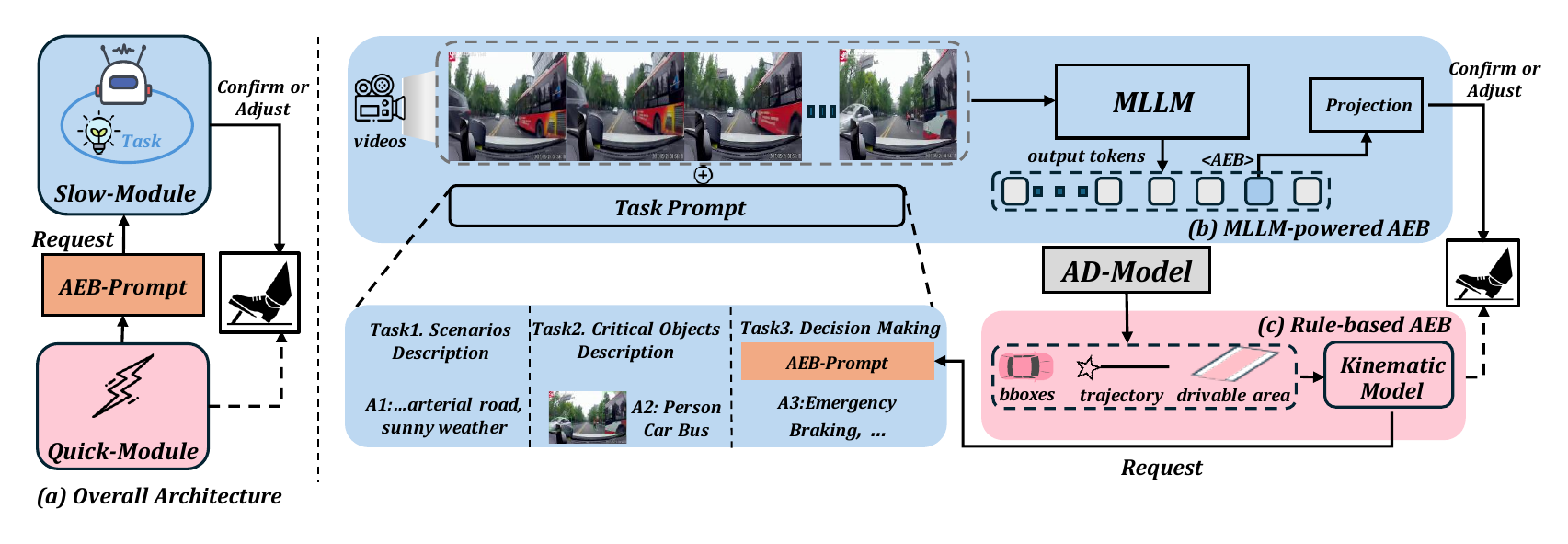}}
\caption{
\textbf{Overview of our method.} Dual-AEB \textbf{(a)} includes both the quick (rule-based AEB) and slow (MLLM-powered AEB) modules. After receiving information from autonomous driving models (AD-Models), the braking signal can either be directly output by \textbf{(c)}, as indicated by dashed lines, or sent to \textbf{(b)}, as indicated by solid lines, where the MLLM-powered AEB evaluates and decides whether to confirm or adjust.
}
\label{fig:mainfig}
\end{figure*}

\subsection{Overview}
The entire workflow of the Dual-AEB system is illustrated in the Fig.~\ref{fig:mainfig}. This system consists of two main components: the quick module, called the rule-based AEB, and the slow module, named the MLLM-powered AEB. The rule-based AEB, using conventional rule-based methods, is responsible for the initial decision. When triggered, the quick module packages this initial decision into text, named as the \textit{AEB-Prompt}, and sends it to the slow module, the MLLM-powered AEB. The slow module, utilizing MLLM to analyze the received information, makes the final decision, either confirming or adjusting the quick module's initial decision. This framework can be seamlessly integrated with other autonomous driving algorithms, making it a flexible framework that can be easily expanded or updated.

\subsection{Rule-Based AEB Module} 
The rule-based AEB module receives perception and planning results from the autonomous driving modules (AD models), including the bounding boxes of agents ({\it B}), the ego vehicle's planned trajectory ({\it P}), and the drivable area ({\it D}). This information is used to evaluate potential collisions within the trajectory horizon ({\it H}) by utilizing the kinematic bicycle model ({\it K}) \cite{bicycle_model}, with a step size ({$\Delta t$}) for temporal progression. This module then uses  Time To Collision ({\it TTC}) and Collision ({\it C}) \cite{navsim} as evaluation metrics to assess the need for emergency braking \cite{Dauner2023CORL}. The decision to initiate braking is then made by comparing the calculated trigger time (\( t_{\text{trigger}} \)) with a predefined threshold (\( t_{\text{threshold}} \)). The detailed steps of this process are outlined in Algorithm~\ref{alg:rule-based}.

\begin{algorithm}
\caption{Rule-Based AEB Module}
\label{alg:rule-based}
\begin{algorithmic}[1]
\State \textbf{Input:} \( \textit{B}_{\textit{ego},t_0}, \textit{B}_{\textit{other},t_0}, \textit{P}, \textit{D}, \textit{K}, \Delta t, \textit{H} \)
\State \textbf{Initialize:} \( \textit{trigger\_times} = [] \)

\For{\( t = \Delta t \) to \( \textit{H} \times \Delta t \) step \( \Delta t \)}
    \State \( \textit{B}_{\textit{ego},t} = \textit{UpdateEgoBBox}(\textit{B}_{\textit{ego},t-\Delta t}, \textit{P}[t/\Delta t], \textit{K}) \)
    \State \( \textit{B}_{\textit{other},t} = \textit{UpdateOtherBBox}(\textit{B}_{\textit{other},t-\Delta t}, \textit{K}) \)
    \State \( \textit{TTC}_t = \textit{CalculateTTC}(\textit{B}_{\textit{ego},t}, \textit{B}_{\textit{other},t}, \textit{D}) \)
    \State \( \textit{C}_t = \textit{CalculateCollision}(\textit{B}_{\textit{ego},t}, \textit{B}_{\textit{other},t}, \textit{D}) \)
    \If{\( \textit{TTC}_t < t_{\textit{threshold}} \) or \( \textit{C}_t \)}
        \State Append \( t \) to \( \textit{trigger\_times} \)
    \EndIf
\EndFor

\State \textbf{Decision:}
\If{\( \textit{trigger\_times} \) is not empty}
    \State \Return Brake
\EndIf

\end{algorithmic}
\end{algorithm}

\subsection{MLLM-Powered AEB Module}
The MLLM-Powered AEB module processes a series of sequential front-view driving images accompanied by predefined task prompts. It consists of a trained MLLM backbone and a projection network. The MLLM backbone analyzes these inputs and generates textual responses, termed Text Generation (TG), which include an \textless AEB\textgreater~token to indicate when emergency braking is needed. The projection network then utilizes the hidden state associated with the \textless AEB\textgreater~token, applying a linear layer followed by a sigmoid activation to output a braking signal—referred to as Braking Signal Generation (BSG)—in the range of 0 to 1.

\subsection{Training Data Construction}
\label{sub:training_data}
The data required for Dual-AEB training includes front-view video inputs, question-answer pairs, and braking signals. We generated this data for the MLLM-powered AEB by utilizing the MM-AU\cite{MM-AU} dataset, which captures real-world accident scenarios, and the Bench2Drive\cite{jia2024bench} dataset, featuring complex simulation scenarios from CARLA\cite{dosovitskiy2017carla}. Videos are directly available from both datasets, and Bench2Drive also provides braking signals. For MM-AU, braking signals are derived by analyzing the chronological sequence of traffic accidents. For instance, in video sequences where a collision is imminent, emergency braking is required, and in such cases, the ground truth for the braking signal is set to 1.

For the question-answer pairs, we designed three sub-tasks focused on driving scenarios, critical objects, and decision-making processes, structuring the reasoning process in a step-by-step manner \cite{wei2023chainofthought, jin2023adapt}. Using GPT-4, we generated diverse question-answer templates and filled them with ground truth. Details are outlined below.

\textbf{Scenarios Description.} The driving environment significantly influences the complexity of driving tasks \cite{tian2024drivevlmconvergenceautonomousdriving}. For instance, rural roads are more likely to encounter sudden appearances of animals. Annotations include details on weather variations, visibility, road traction, time of day, and road types. An example of an annotated scenario description question-answer pair is, \textbf{Q}: ``{\it What are the environmental details captured in this driving video?}" \textbf{A}: ``{\it The ego vehicle navigates an arterial roadway under clear, sunny conditions in an urban environment during daylight.}"

\textbf{Critical Objects Description.} Annotations for critical objects include details such as 2D bounding boxes in $(x\_min, y\_min, x\_max, y\_max)$ format, distance to the ego vehicle, traffic signals, and the intentions of surrounding agents. For example, ``{\it A black vehicle with a blinking left turn signal, located at [(197, 474), (295, 667)], is 14.26 meters away from the ego vehicle, suggesting it is preparing to make a left turn.}" These details provide critical information for the system to determine the precise position of the object within the scene. With this information, Dual-AEB can forecast the object's future movements, resulting in more accurate and informed decision-making in various scenarios.

\textbf{Decision Making.} In the final sub-task, questions are supplemented with {\it AEB-Prompt}, which represent initial decisions generated by the rule-based AEB module. For example, a {\it AEB-Prompt} might be, ``{\it Initial decision: A collision with the black vehicle on the left is expected in 1.2 seconds, and I decide to brake.}" To prevent the MLLM-powered AEB from becoming overly reliant on {\it AEB-Prompt}, ${50\%}$ of the training data includes incorrect initial decisions. We categorize AEB actions into three meta-actions: {\it Normal}, {\it Early Warning}, and {\it Emergency Braking} \cite{accidents_4}. A possible answer might be, ``{\it Early Warning. The presence of a truck and another black car ahead requires heightened awareness. \textless AEB\textgreater}"

\subsection{Training and Inference}
During training, AD models are trained on Bench2Drive\cite{jia2024bench}, and the MLLM-powered AEB employs a composite loss function that integrates both TG and BSG to ensure cohesive semantic learning. The overall loss \(\mathcal{L}\) is defined as:
\begin{equation}
\begin{aligned}
\mathcal{L} = & -\sum_{n=1}^N \sum_{i=1}^V y_{n,i} \log(\hat{y}_{n,i}) \\
              & - \left[ z \log(\hat{z}) + (1 - z) \log(1 - \hat{z}) \right],
\end{aligned}
\end{equation}
where \(\hat{y}_{n,i}\) and \(y_{n,i}\) represent the predicted and ground truth probabilities of word \(i\) at position \(n\) in the text sequence, with \(V\) being the vocabulary size. The variables \(z\) and \(\hat{z}\) denote the ground truth and predicted braking signals, respectively.

During inference, the rule-based AEB receives perception and planning results from the pre-trained AD model. It can either directly output its braking signal or initiate an interaction with the MLLM-powered AEB via the \textit{AEB-Prompt}. If interaction is triggered, the MLLM-powered AEB engages in a multi-round Q\&A process to address the three sub-tasks. Ultimately, during the decision-making task, it determines whether to confirm or adjust the initial decision made by the rule-based AEB. The final braking signal is then output after passing through the projection network.

\section{Experiments}

\subsection{Datasets}
To provide a more comprehensive evaluation of our method, we assess it on two datasets. We perform open-loop evaluations using both MM-AU\cite{MM-AU} and Bench2Drive\cite{jia2024bench}, and conduct closed-loop evaluations using the Bench2Drive\cite{jia2024bench} benchmark. We carefully construct 120,113 samples from MM-AU and 132,922 samples from Bench2Drive, with the proportions of {\it Normal}, {\it Early Warning}, and {\it Emergency Braking} approximately at 1:1:1. Using a 9:1 ratio, we split this data into training and test sets.

\subsection{Metrics}
In the open-loop evaluation, we focus on the accuracy of the model's predicted braking signals and the quality of the generated text. For the Braking Signal Generation (BSG), we utilize standard AEB task metrics\cite{learning_method_6}, namely {\it Precision} and {\it Recall}. Positive cases refer to situations where AEB activation is necessary, and negative cases to those where it is not needed. True Positives (TP) are instances of correctly triggered {\it Emergency Braking}, while True Negatives (TN) are correct non-activations ({\it Early Warning/Normal}). Conversely, False Positives (FP) represent erroneous activations, and False Negatives (FN) denote instances where necessary AEB actions are missed. The formulas for {\it Precision} and {\it Recall} are provided as follows: 
\begin{equation}
\label{metrics}
    \text{Precision} = \frac{TP}{TP + FP}, ~ \text{Recall} = \frac{TP}{TP + FN},
\end{equation}
for the Text Generation (TG) aspect, we evaluate the quality of the generated text using BLEU4\cite{papineni2002bleu}, METEOR\cite{banerjee2005meteor}, and ROUGE-L\cite{rougel} metrics.

In the closed-loop evaluation, our primary focus is on the model's overall driving performance. We utilize the {\it Driving Score} and {\it Success Rate} metrics from Bench2Drive \cite{jia2024bench}. The {\it Driving Score} aggregates multiple driving metrics from Bench2Drive into a weighted sum, while the {\it Success Rate} measures the proportion of successfully completed scenarios out of the total number of scenarios. Additionally, we introduced the {\it Collision Rate}, defined as the average number of collisions per scenario, to better assess the model's ability to avoid collisions.
\subsection{Implementation Details}

For the rule-based AEB module, we instantiate two widely used models, UniAD \cite{hu2023_uniad} and VAD \cite{jiang2023vad}, to generate perception and planning results, thereby validating the flexibility of our framework. We adopt a step size of 0.2 seconds and a 3-second horizon, aligning with the 3-second future trajectory provided by the Bench2Drive benchmark planner. For the MLLM-powered AEB module, we employ the state-of-the-art LLaVA-OneVision \cite{li2024llavaonevisioneasyvisualtask} model as the backbone and perform full fine-tuning of all its components. Following the recommendations in \cite{li2024llavaonevisioneasyvisualtask}, a learning rate of \(2 \times 10^{-6}\) is applied to the vision encoder, while a learning rate of \(1 \times 10^{-5}\) is used for the other components.

For closed-loop evaluation, the rule-based AEB module is provided with information from VAD or UniAD and interacts with the MLLM-powered AEB module every 2.5 seconds, and the MLLM-powered AEB is trained on the Bench2Drive dataset before being evaluated in CARLA. 

All experiments are conducted on a server equipped with 8 NVIDIA A100 80G GPUs. To ensure the deployability of the framework in real-world scenarios, inference time consumption tests are performed on consumer-grade devices; here, we use an NVIDIA Jetson Orin.

\subsection{Closed-Loop Experimental Results}
We leverage Qwen-0.5B as the foundational language model for LLaVA-OneVision to ensure faster response times within the Bench2Drive closed-loop simulation benchmark. Detailed results are provided in Table~\ref{tab:closeloop_result}. Specifically, with Dual-AEB, VAD's {\it Driving Score} increased by $\textbf{14.74\%}$, while the {\it Success Rate} remained unchanged. For UniAD-Base, the {\it Driving Score} improved by $\textbf{6.84\%}$, and the {\it Success Rate} increased from 9.54 to 10.00. Among all evaluated models \cite{zhai2023rethinkingopenloopevaluationendtoend, hu2023_uniad, jiang2023vad, wu2022trajectoryguidedcontrolpredictionendtoend, jia2023thinktwicedrivingscalable, jia2023driveadapterbreakingcouplingbarrier}, VAD with Dual-AEB achieved the highest performance in terms of {\it Driving Score}.

Dual-AEB helps improve the closed-loop performance of these models by providing effective braking decisions. \textbf{However, since it does not alter the autonomous driving model's output trajectory, the improvement on {\it Success Rate} is limited.} Despite this, the overall driving performance of these end-to-end models is enhanced, demonstrating the potential for Dual-AEB to be integrated into other autonomous driving systems.

\begin{table}[h!]
\caption{Closed-loop results in Bench2Drive. \textsuperscript{*} represents expert feature distillation. The metrics include Driving Score (\textbf{D}), Success Rate (\textbf{S}) and the improvement of success rate ({\textbf{${\Delta}$}S}).
}
\centering
\setlength{\tabcolsep}{2pt} 
\begin{tabular}{c|c|c|S[table-format=2.2]|c}
\toprule
\multirow{2}{*}{\textbf{Method}} & \multirow{2}{*}{\textbf{Input}} & \multicolumn{3}{c}{\textbf{Metrics ↑}} \\
\cmidrule(lr){3-5}
 & & {\textbf{D}} & {\textbf{S}} & {\textbf{${\Delta}$}S}\\
\midrule
TCP*\cite{wu2022trajectoryguidedcontrolpredictionendtoend} & {Ego State + Front Cameras} & 23.63 & \phantom{0}\textcolor{gray}{7.72} & - \\
TCP-ctrl* & {Ego State + Front Cameras} & 18.63 & \phantom{0}\textcolor{gray}{5.45} & - \\
TCP-traj* & {Ego State + Front Cameras} & 36.78 & \textcolor{gray}{26.82} & - \\
ThinkTwice*\cite{jia2023thinktwicedrivingscalable} & {Ego State + 6 Cameras} & 39.88 & \textcolor{gray}{28.14} & - \\
DriveAdapter*\cite{jia2023driveadapterbreakingcouplingbarrier} & {Ego State + 6 Cameras} & 42.91 & \textcolor{gray}{30.71} & - \\
\midrule
AD-MLP\cite{zhai2023rethinkingopenloopevaluationendtoend} & {Ego State} & \phantom{0}9.14 & \phantom{0}\textcolor{gray}{0.00} & - \\
UniAD-Tiny\cite{hu2023_uniad} & {Ego State + 6 Cameras} & 32.00 & \phantom{0}\textcolor{gray}{9.54} & - \\
UniAD-Base\cite{hu2023_uniad} & {Ego State + 6 Cameras} & 37.72 & 9.54 & - \\
VAD\cite{jiang2023vad} & {Ego State + 6 Cameras} & 39.42 & 10.00 & - \\
\midrule
\textbf{UniAD-Base+Dual-AEB} & {Ego State + 6 Cameras} & 40.32 & 10.00 & \textbf{0.06} \\
\textbf{VAD+Dual-AEB} & {Ego State + 6 Cameras} & \textbf{45.23} & 10.00 & 0.00 \\
\bottomrule
\end{tabular}
\label{tab:closeloop_result}
\end{table}
\subsection{Open-Loop Experimental Results}

\begin{figure}[thpb]
\centerline{\includegraphics[width=0.5\textwidth]{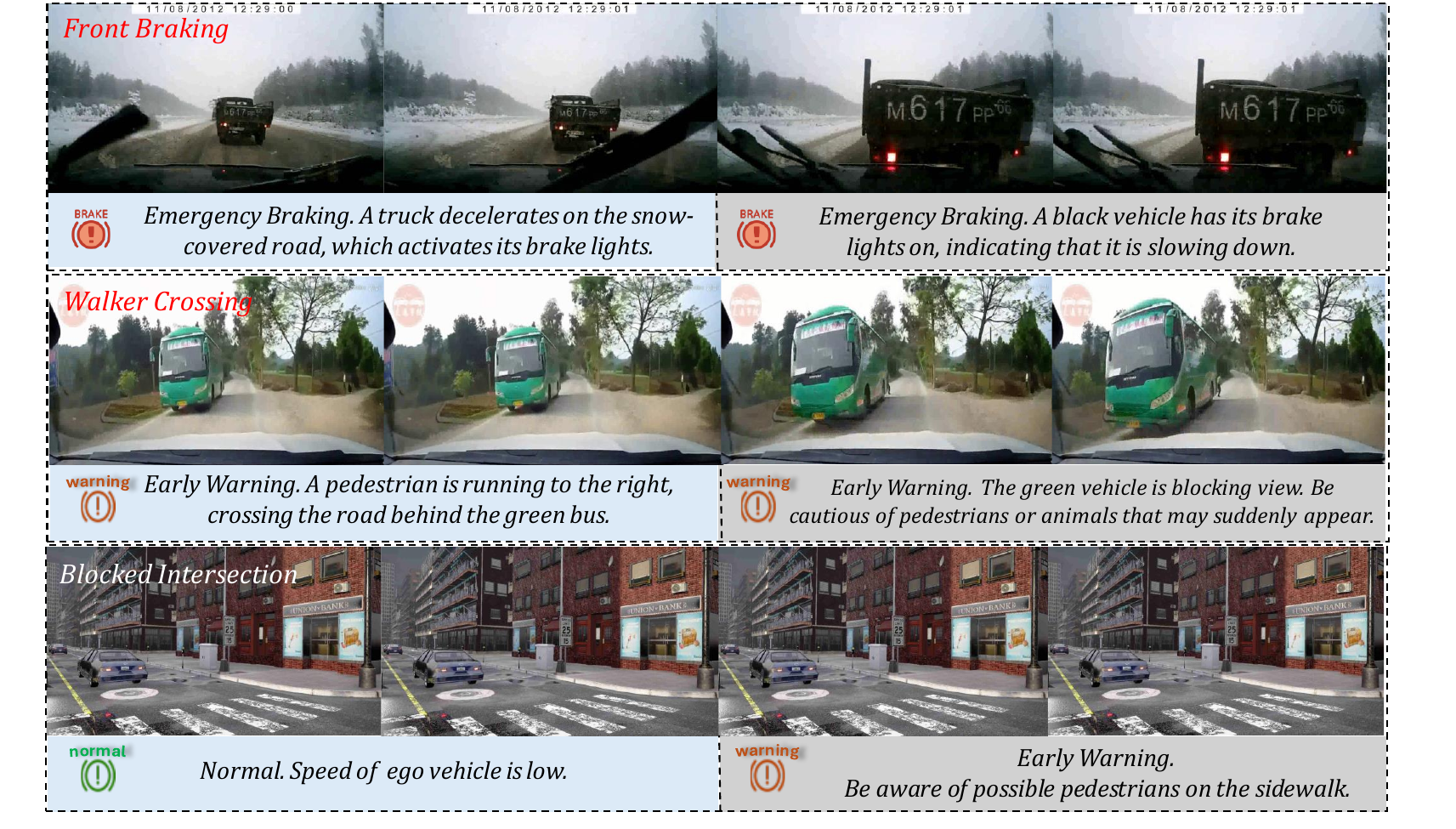}}
\caption{Qualitative analysis: MLLM-powered AEB provided reasonable descriptions for different meta actions. The left side (\textcolor{opengt}{\rule{3mm}{3mm}}) represents the ground truth results, while the right side (\textcolor{openpred}{\rule{3mm}{3mm}}) represents the predicted results.}
\label{fig:OpenLoop}
\end{figure}

In this part, we aim to evaluate the comprehensive scene understanding capabilities of MLLM-powered AEB in real-world scenarios. We compare the {\it Precision}, {\it Recall}, and text generation quality of the trained LLaVA-OneVision model based on Qwen-0.5B and Qwen-7B, taking YOLO-AEB \cite{Wu2021RealTimeVA} as the baseline. Since the YOLO model processes single images, we take the last frame from each video split as input. We present the quantification results in Table~\ref{tab:openloop_comparison}, and provide qualitative examples in Fig.~\ref{fig:OpenLoop}.

Compared to YOLO-AEB, MLLM-powered AEB achieves higher {\it Precision} and {\it Recall}, thanks to the powerful and comprehensive scene understanding capabilities of MLLMs, which enable more effective braking decisions. Additionally, the increase in model scale further contributes to the observed performance improvements.

On the Bench2Drive benchmark, both YOLO and the MLLM-powered AEB demonstrate improved performance compared to their results on MM-AU. This enhancement is likely due to the simpler and more uniform objects in the simulated scenarios, which facilitate the models in learning effective braking behaviors. However, these improvements do not extend to {\it Precision}. Specifically, the MLLM's performance is somewhat hindered because Bench2Drive's simulation data is generated by an expert model \cite{li2024think} that consistently avoids danger by braking. As a result, some scenarios appear inherently safe, reducing the model's tendency to initiate braking and leading to lower {\it Precision}.

\begin{table}[h!]
\caption{Impact of different methods in open-loop evaluations (\textbf{P}: Precision, \textbf{Recall}, \textbf{B4}: BLEU-4, \textbf{M}: METEOR, \textbf{R}: ROUGE-L). The models evaluated include Qwen-0.5B (Q0.5B), Qwen-7B (Q7B), and YOLO, across two datasets: MM-AU and B2D (Bench2Drive).}
\centering
\begin{tabular}{cc|ccccc}
\toprule
\multirow{2}{*}{\textbf{Dataset}} & \multirow{2}{*}{\textbf{Method}} & \multicolumn{5}{c}{\textbf{Metrics ↑}} \\
\cmidrule{3-7}
& & \textbf{P} & \textbf{Recall} & \textbf{B4} & \textbf{M} & \textbf{R} \\ 
\midrule
\multirow{3}{*}{MM-AU} & YOLO & 26.62 & 38.99 & - & - & - \\ 
& Q0.5B & 72.90 & 73.80 & 13.18 & 30.86 & 33.69 \\ 
& Q~~7B & \textbf{73.60} & \textbf{75.20} & \textbf{20.12} & \textbf{36.30} & \textbf{37.95} \\ 
\midrule
\multirow{3}{*}{B2D} & YOLO & 56.27 & 67.34 & - & - & - \\ 
& Q0.5B & 65.25 & \textbf{94.35} & 21.74 & 42.74 & 40.04 \\ 
& Q~~7B & \textbf{67.36} & 93.14 & \textbf{28.09} & \textbf{46.85} & \textbf{45.07} \\ 

\bottomrule
\end{tabular}
\label{tab:openloop_comparison}
\end{table}

\subsection{Ablation Study}

\textbf{Impact of Different AEB Modules.} To assess the effectiveness of various Dual-AEB modules in providing accurate braking decisions, we conducte an evaluation. The specific results are presented in Table~\ref{tab:closeloop_comparison}. 

After integrating the rule-based AEB module into VAD and UniAD, both models exhibit a significant reduction in {\it Collision Rate}, resulting in an improved {\it Driving Score}. Replacing the rule-based AEB with an MLLM-powered AEB, which possesses advanced scene understanding capabilities, further decreases the {\it Collision Rate}, highlighting the MLLM's stronger ability to perceive potential dangers. Finally, with the implementation of the full Dual-AEB, the {\it Driving Score} improves once more. This demonstrates that the rule-based AEB can provide rapid responses in dangerous scenarios when the MLLM is not invoked—since it operates intermittently—thereby compensating for the limitations of the MLLM.

\textbf{Impact of Different Task Modules.} In previous experiments, we design three sequential tasks to guide the model's TG process. This section compares the performance of different task combinations on the MM-AU and Bench2Drive open-loop datasets (Table~\ref{tab:task_comparison}). The results indicate that incorporating additional tasks enhances the inference performance of the MLLM-powered AEB module, with the critical objects description task providing the most substantial improvement. This task allows the model to better understand the intentions of other objects, resulting in more effective braking.

\begin{table}[h!]
\caption{Impact of different AEB modules in close-loop evaluation (\textbf{R}: Rule-based AEB, \textbf{M}: MLLM-powered AEB). \\
The metrics include Driving Score (\textbf{D}), Success Rate (\textbf{S}), and Collision Rate (\textbf{C}).}
\centering
\setlength{\tabcolsep}{12pt} 
\begin{tabular}{c|cc|ccc}
\toprule
\multirow{2}{*}{\textbf{Models}} & \multicolumn{2}{c|}{\textbf{Modules}} & \multicolumn{3}{c}{\textbf{Metrics}} \\
\cmidrule(lr){2-3} \cmidrule(lr){4-6}
 & \textbf{R} & \textbf{M} & \textbf{D ↑} & \textbf{S ↑} & \textbf{C ↓} \\
\midrule
\multirow{4}{*}{VAD} 
 & \ding{55} & \ding{55} & 39.42 & 10.00 & 70.19 \\
 & \ding{52} & \ding{55} & 43.22 & 10.00 & 56.43 \\
 & \ding{55} & \ding{52} & 44.75 & 10.00 & 53.42 \\
 & \ding{52} & \ding{52} & \textbf{45.23} & \textbf{10.00} & \textbf{50.86} \\
\midrule
\multirow{4}{*}{UniAD} 
 & \ding{55} & \ding{55} & 37.72 & 9.54 & 87.06 \\
 & \ding{52} & \ding{55} & 38.47 & 9.54 & 74.53 \\
 & \ding{55} & \ding{52} & 39.46 & 10.00 & 71.68 \\
 & \ding{52} & \ding{52} & \textbf{40.32} & \textbf{10.00} & \textbf{69.02} \\
\bottomrule
\end{tabular}
\label{tab:closeloop_comparison}
\end{table}

\begin{table}[h!]
\centering
\caption{Impact of different task modules (\textbf{S}: Scenarios Description, \textbf{O}: Critical Objects Description, \textbf{D}: Decision Making).}
\label{tab:task_comparison}
\setlength{\tabcolsep}{10pt}
\begin{tabular}{c|ccc|cc}
\toprule
\textbf{Datasets} & \textbf{S} & \textbf{O} & \textbf{D} & \textbf{Precision} & \textbf{Recall} \\
\midrule
\multirow{4}{*}{MM-AU} 
& \ding{55} & \ding{55} & \ding{55} & 56.80 & 57.62 \\
& \ding{52} & \ding{55} & \ding{55} & 58.68 & 62.54 \\
& \ding{52} & \ding{52} & \ding{55} & 71.44 & 68.60 \\
& \ding{52} & \ding{52} & \ding{52} & \textbf{72.90} & \textbf{73.80} \\
\midrule
\multirow{4}{*}{Bench2Drive} 
& \ding{55} & \ding{55} & \ding{55} & 57.44 & 68.32 \\
& \ding{52} & \ding{55} & \ding{55} & 59.14 & 72.69 \\
& \ding{52} & \ding{52} & \ding{55} & 62.53 & 88.96 \\
& \ding{52} & \ding{52} & \ding{52} & \textbf{65.25} & \textbf{94.35} \\
\bottomrule
\end{tabular}
\end{table}

\textbf{Impact of Different Trigger Intervals of the MLLM-powered AEB.} In previous experiments, we test the MLLM-powered AEB module in Bench2Drive with a 2.5-second trigger interval. Here, we evaluate the impact of different intervals: 0.5, 2.5, and 5 seconds. As shown in Table~\ref{tab:trigger_time_comparison}, shorter intervals improve {\it Driving Score}, but the 0.5-second interval provides only marginal gains and significantly increases inference time. Thus, a 2.5-second interval offers a better balance between performance and efficiency.

\textbf{Comparison of Inference Time Consumption on Different Devices.} After conducting frequency analysis, we perform average inference time tests on the NVIDIA Jetson Orin and optimize the models using TensorRT. The results are presented in Table~\ref{tab:latency_exp}. To achieve faster response times, we develop a version of Dual-AEB that executes only the decision-making task (Dual-AEB-S), based on the original Dual-AEB that executes all tasks (Dual-AEB-F). By combining trigger time with accelerated inference, we further reduce the average inference time.


\subsection{Cases Study}

\begin{figure}[thpb]
\centerline{\includegraphics[width=0.5\textwidth]{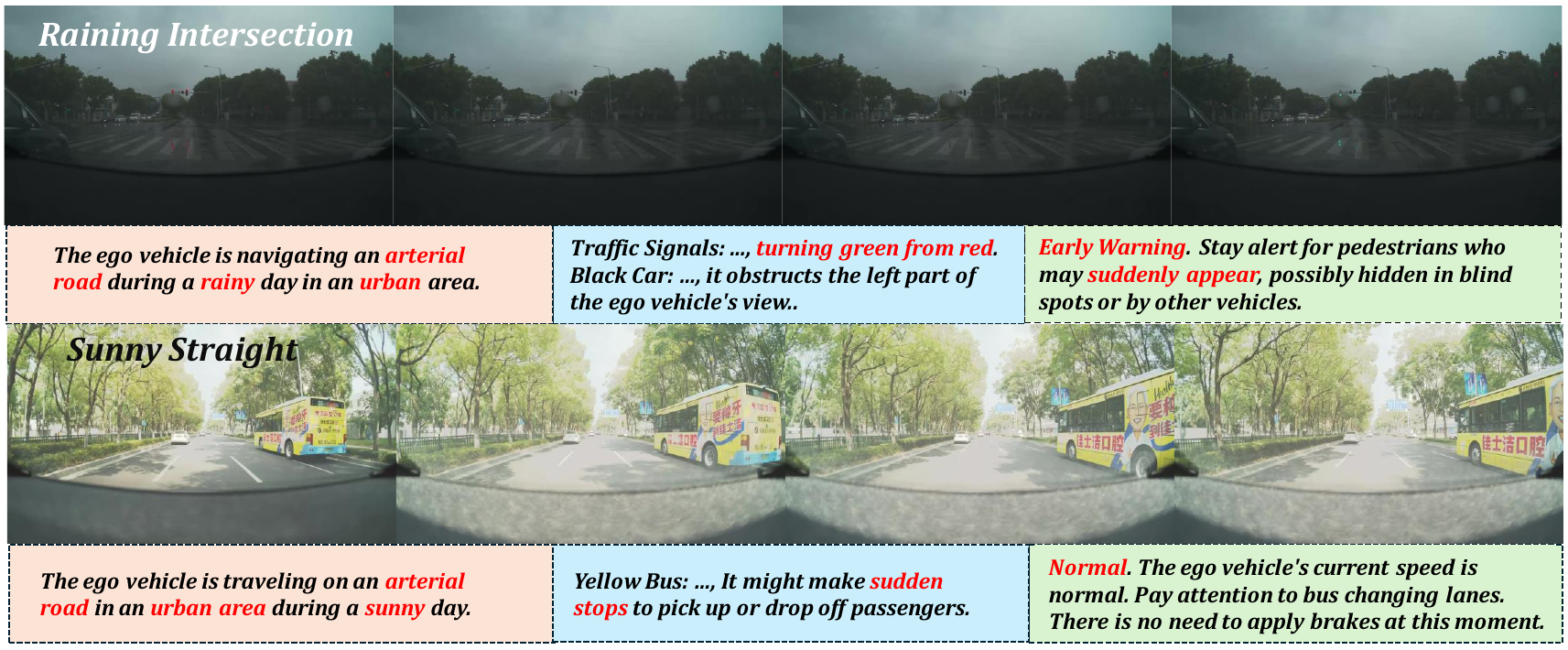}}
\caption{Qualitative analysis: Dual-AEB provides reasonable descriptions in our in-house dataset. \textcolor{scene}{\rule{3mm}{3mm}} represents the scenarios description task, \textcolor{object}{\rule{3mm}{3mm}} represents the objects description task, and \textcolor{action}{\rule{3mm}{3mm}} represents the decision making task.}
\label{fig:inhouse}
\vspace{-0.5cm}
\end{figure}

MM-AU comes from network dashcam recordings, and Bench2Drive originates from simulations. There are still some differences between them and real-world driving data. To further explore the capabilities of our model in real-world driving scenarios, we conduct tests on our in-house dataset, which is collected from a vehicle equipped with a complete autonomous driving hardware and software system. Fig.~\ref{fig:inhouse} shows several case results, which are similar to the scenarios mentioned in Fig.~\ref{fig:teaser}.

\begin{table}[h!]
\caption{Impact of different trigger intervals on Bench2Drive (\textbf{D}: Driving Score, \textbf{C}: Collision Rate, \textbf{T}: Average Inference Time). }
\centering
\setlength{\tabcolsep}{10pt} 
\begin{tabular}{ccccc}
\toprule
\textbf{Models} & \textbf{Trigger Time} & \textbf{D ↑} & \textbf{C ↓} & \textbf{T (s) ↓} \\
\midrule
\multirow{3}{*}{VAD} 
 & 0.5 & \textbf{45.92} & \textbf{49.98} & 8.10 \\
 & 2.5 & 45.23 & 50.86 & 1.55 \\
 & 5.0 & 42.14 & 54.23 & \textbf{0.80} \\
\midrule
\multirow{3}{*}{UniAD} 
 & 0.5 & \textbf{40.56} & \textbf{68.74} & 8.25 \\
 & 2.5 & 40.32 & 69.02 & 1.60 \\
 & 5.0 & 38.87 & 73.54 & \textbf{0.80} \\
\bottomrule
\end{tabular}
\label{tab:trigger_time_comparison}
\end{table}

\begin{table}[ht]
\centering
\caption{Comparison of average inference time consumption results on different devices using TensorRT.}
\setlength{\tabcolsep}{12pt} 
\begin{tabular}{c|c|S[table-format=2.2]S[table-format=2.2]} 
\toprule
\multirow{2}{*}{\textbf{Modules}} & \multirow{2}{*}{\textbf{TensorRT}} & \multicolumn{2}{c}{\textbf{Devices}} \\
\cmidrule(lr){3-4}
& & \textbf{A100} & \textbf{Jetson Orin} \\
\midrule
\multirow{2}{*}{Dual-AEB-F} 
& \ding{55} & 1.55s & 11.44s \\
& \ding{52} & 0.96s & 4.86s \\
\midrule
\multirow{2}{*}{Dual-AEB-S} 
& \ding{55} & 0.68s & 3.87s \\
& \ding{52} & 0.31s & 1.21s \\
\bottomrule
\end{tabular}
\label{tab:latency_exp}
\end{table}

In the first scenario, conventional AEB often proceed without providing an early warning to the driver, which may result in insufficient reaction time when a pedestrian suddenly appears. In the second scenario, conventional AEB may mistakenly identify an advertisement on a bus as a pedestrian, triggering unnecessary emergency braking. In contrast, Dual-AEB offers early warnings and filters out false positives, thereby enhancing both safety and the overall effectiveness of the system.

\section{Conclusion}

In this study, we present Dual-AEB, an innovative method that integrates conventional rule-based modules with advanced Multimodal Large Language Models (MLLMs) to enhance autonomous emergency braking (AEB). The conventional component ensures rapid initial responses, while the MLLM component enhances decision accuracy by processing complex environmental data and ego-vehicle states. Evaluations on open-loop and closed-loop benchmarks demonstrate the effectiveness of this system in providing robust braking strategies. Further testing on real-world scenarios confirms the practical applicability and strong performance of the Dual-AEB system. Enhancing precision in real-world applications is the focus of our future work.


\bibliographystyle{IEEEtran}
\bibliography{References}

\end{document}